%% file: acl_latex.tex
\documentclass[11pt]{article}

\usepackage[preprint]{acl}

\usepackage{times}
\usepackage{latexsym}

\usepackage[T1]{fontenc}

\usepackage[utf8]{inputenc}

\usepackage{microtype}

\usepackage{inconsolata}

\usepackage{graphicx}

\usepackage{amsmath}
\usepackage{listings}

\usepackage{listings}
\usepackage{xcolor}

\usepackage{booktabs}
\usepackage{tabularx}

\colorlet{punct}{red!60!black}
\definecolor{background}{HTML}{FFFFFF}
\definecolor{delim}{RGB}{20,105,176}
\colorlet{numb}{magenta!60!black}

\lstdefinelanguage{json}{
    basicstyle=\normalfont\ttfamily,
    numberstyle=\scriptsize,
    stepnumber=1,
    numbersep=8pt,
    showstringspaces=false,
    breaklines=true,
    backgroundcolor=\color{background},
}

\title{Reasoning or Rationalization? The Role of Justifications in Masked Diffusion Models for Fact Verification}

\author{Jacob Devasier \\
  The University of Texas at Arlington \\
  \texttt{jacob.devasier@uta.edu} 
  }

\begin{document}
\maketitle

\begin{abstract}
Unlike autoregressive models, which generate tokens sequentially and benefit from reasoning-before-answering strategies such as Chain-of-Thought, Masked Diffusion Language Models (MDLMs) refine all sequence positions simultaneously, raising questions about how these models handle tasks requiring justified verdicts. In this work, we investigate the dynamics of MDLM reasoning on fact verification, examining whether justifications serve as genuine reasoning or post-hoc rationalization. We observe that MDLMs typically converge on a verdict early in the diffusion process, treating it as a global anchor that is resolved before the justification is complete. Crucially, enforcing a reasoning-first constraint via delayed verdict unmasking actively degrades performance, dropping accuracy from 86.2\% to 71.9\% as accumulating justification tokens introduce inconsistencies that override initially correct predictions. Interventional experiments reveal that the model rationalizes incorrect forced verdicts in 56\% of cases, and that verdicts are strongly causally dependent on justification quality (57.3\% accuracy with corrupted justifications vs.\ 97.1\% with ground-truth). This causal dependence explains the degradation under forced deliberation: as the model generates noisy justification tokens, it conditions on them, gradually overriding its initially correct assessment. Our findings suggest that for fact verification with MDLMs, extended deliberation can be counterproductive, risking the dilution of accurate early predictions with noise introduced during justification generation.
\end{abstract}

\input{sections/1-intro}

\input{sections/3-methodology}

\input{sections/4-experiments}

\input{sections/5-conclusion}

\section*{Limitations}
First, while AVeriTeC involves real-world claims that are more complex than Wikipedia-based benchmarks like FEVER, the inclusion of gold-standard evidence in the prompt reduces task complexity. It remains an open question whether refinement drift persists in domains requiring strict, multi-step deduction, such as advanced mathematics (e.g., AIME), where intermediate steps are logical prerequisites for the solution rather than linguistic rationalizations of an intuitive verdict.

Second, our experiments are limited to LLaDA-8B, the only high-performing MDLM currently available at this scale. The susceptibility to refinement drift may be a function of model scale, pretraining data, or optimization procedure; larger models with greater capacity for global attention may maintain coherence over longer reasoning traces.

Third, our findings may not generalize beyond masked diffusion models. While the decoding process shares similarities, it is unclear whether discrete diffusion models such as SEDD~\cite{lou2024discretediffusionmodelingestimating} or continuous diffusion models~\cite{li2022diffusionlmimprovescontrollabletext} would exhibit the same behavior.


\bibliography{custom}

\appendix

\input{sections/appendix}

\end{document}

%% file: sections/1-intro.tex
\section{Introduction}
Traditional autoregressive large language models are built on the next-token prediction objective~\cite{vaswani-attention}, which necessitates a causal, sequential approach to text generation. In this framework, Chain-of-Thought prompting~\cite{wei2023-chain} is often viewed as a causal prerequisite for complex reasoning, where intermediate steps provide the necessary context for a final answer~\cite{lanham2023measuringfaithfulnesschainofthoughtreasoning,turpin2023languagemodelsdontsay}. Conversely, Masked Diffusion Language Models (MDLMs)~\cite{sahoo2024simpleeffectivemaskeddiffusion}, such as LLaDA~\cite{nie2025largelanguagediffusionmodels}, model the conditional distribution of all sequence positions simultaneously. While MDLMs provide significant flexibility in decoding order, recent evaluations~\cite{horvitz2025computeleftbehindrethinking} on math and coding benchmarks suggest that any-order decoding algorithms often underperform or perform similarly to standard left-to-right sampling. This observation raises questions about how to effectively leverage the bidirectional capabilities of MDLMs, particularly for tasks requiring justified decisions.

In this paper, we investigate how iterative refinement affects the performance and consistency of MDLM-generated reasoning traces in fact verification. We evaluate LLaDA-8B~\cite{nie2025largelanguagediffusionmodels} on the AVeriTeC dataset~\cite{schlichtkrull2023averitecdatasetrealworldclaim}, a benchmark for real-world claim verification featuring diverse claims that are less likely to appear in pretraining data than Wikipedia-based sources. Fact verification provides a particularly interesting testbed for studying MDLM behavior: unlike mathematical reasoning where intermediate steps are strict logical prerequisites, fact-checking justifications may represent either genuine evidence synthesis or post-hoc rationalization of intuition.

\input{figures/example}

We begin by investigating a fundamental difference between causal and non-causal architectures: the sensitivity to output ordering. In autoregressive models, generating the justification before the verdict is critical, as it allows the model to condition its final prediction on the generated reasoning traces. Reversing this order often forces the model to commit to a verdict prematurely, degrading performance~\cite{pelrine-etal-2023-towards}. We investigate whether MDLMs, with their bidirectional attention, are susceptible to this same limitation. Our results show that LLaDA-8B is remarkably robust: it achieves high accuracy regardless of whether the justification precedes or follows the verdict, outperforming the standard LLaMA 3.1 (8B)~\cite{grattafiori2024llama3herdmodels} baseline and matching the performance of Qwen3-8B~\cite{yang2025qwen3technicalreport}. Crucially, however, our analysis reveals that even when explicitly prompted to generate justifications first, LLaDA consistently predicts the verdict within the first few diffusion steps.

This observation raises a critical question: does the subsequent refinement process improve the verdict decision, or does it merely generate supporting justification? To investigate, we constrain the decoding process by preventing the model from predicting a verdict until a specified percentage of the justification has been generated. Rather than improving accuracy, this forced deliberation proves counterproductive: as the deliberation threshold increases from 0\% to 90\%, accuracy steadily degrades from 86.2\% to 71.9\%. By monitoring the evolution of the verdict token across diffusion steps, we find that the model often identifies the correct verdict early in the process but subsequently reverses its prediction as justification tokens accumulate. We refer to this phenomenon as \textit{refinement drift}: the progressive introduction of local inconsistencies through justification generation that conflict with and ultimately override the model's initially correct global assessment.

To further probe the causal relationship between justifications and verdicts, we employ a two-way interventional test that exploits the MDLM's ability to condition on arbitrary subsets of tokens. First, we force the model to generate justifications for incorrect verdicts. We find that the model maintains logical integrity in only 44\% of cases---generating a justification that contradicts the forced verdict and supports the correct answer---while in the remaining 56\%, the model rationalizes the incorrect verdict through logical errors (37\%), factual hallucinations (13\%), or other failures (6\%). Second, we test the model's reliance on its own generated traces by forcing it to predict the verdict conditioned on these corrupted justifications. In this setting, accuracy drops sharply to 57.3\%, driven primarily by a near-total failure on supported claims (17.2\% accuracy). In contrast, when the model is provided with high-quality ground-truth justifications from the AVeriTeC dataset, accuracy reaches 97.1\%. Together, these results demonstrate that LLaDA's verdicts are strongly causally dependent on justification quality, and that this dependence is the mechanism underlying refinement drift: as the model generates noisy justification tokens, it conditions on them, gradually overriding its own initially correct assessment.

%% file: figures/example.tex
\begin{figure*}
  \centering
  \includegraphics[width=0.8\linewidth]{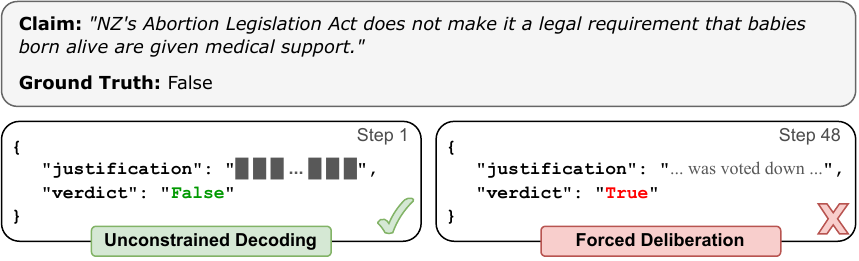}
  \caption{\label{fig:example}An example of refinement drift. When forced to generate most of the justification (regardless of verdict confidence) before predicting the verdict, the model reaches the wrong conclusion.}
  \vspace{-2mm}
\end{figure*}

%% file: sections/3-methodology.tex
\section{Methodology}

We evaluate on the AVeriTeC dataset~\cite{schlichtkrull2023averitecdatasetrealworldclaim}, a benchmark of real-world factual claims paired with verdicts, supporting evidence (in the form of question--answer pairs), and human-annotated justifications. Unlike Wikipedia-centric datasets such as FEVER, AVeriTeC draws claims from diverse web sources, requiring models to synthesize complex and often conflicting evidence. We focus on the core fact verification task: given a claim and its evidence, the model must produce both a verdict (Supported or Refuted) and a justification. This formulation is particularly suited to studying MDLM behavior, as it exposes the tension between the model's global commitment to a verdict and the justification's local adherence to specific facts.

\subsection{Model Selection}
Our study centers on LLaDA-8B~\cite{nie2025largelanguagediffusionmodels}, the first MDLM scaled to a level comparable to modern pretrained autoregressive LLMs and currently the strongest-performing model in this class. Unlike autoregressive LLMs, which employ causal masking to predict tokens sequentially, LLaDA uses a non-causal architecture in which every token attends to every other token. This bidirectional attention allows the model to refine global context (e.g., the verdict) concurrently with local details (e.g., the justification), treating the entire output as a joint distribution. Following prior recommendations~\cite{horvitz2025computeleftbehindrethinking,nie2025largelanguagediffusionmodels}, we unmask one token at a time during decoding.

\subsection{Output Format and Generation Length}
We instruct the model to return its predictions in JSON format to facilitate parsing of the verdict and justification fields. We fix the generation length at 64 tokens, leaving approximately 48 tokens for the justification after accounting for JSON structure and the verdict token. This length was chosen based on preliminary experiments showing that longer outputs frequently led to verdict leakage in the justification (e.g., ``this claim is false because~...''), as it confounds our analysis of how the verdict evolves during the diffusion process.

%% file: sections/4-experiments.tex
\section{Experiments}
We evaluate on the AVeriTeC development set ($N=500$), framing the task as binary classification: Supported vs.\ Refuted. Because ``Refuted'' spans multiple tokens in LLaDA's vocabulary, we map each label to a single-token proxy: ``True'' for Supported and ``False'' for Refuted. Evidence is then provided as question--answer pairs.

\subsection{Baseline Models}
Our primary model is LLaDA-8B\footnote{\url{https://huggingface.co/GSAI-ML/LLaDA-8B-Instruct}}. We compare it against two baselines: LLaMA 3.1 8B\footnote{\url{https://huggingface.co/meta-llama/Llama-3.1-8B-Instruct}}, a standard autoregressive model, and Qwen3-8B\footnote{\url{https://huggingface.co/Qwen/Qwen3-8B}} with extended thinking mode enabled, which serves as a reasoning-augmented baseline. Together, these models allow us to compare LLaDA's any-order generation relative to both standard and reasoning-based autoregressive decoding.

\subsection{Justification Ordering}\label{sec:justification-ordering}
Generating reasoning traces before answers (e.g., Chain-of-Thought prompting) benefits autoregressive LLMs and has been shown to improve fact-checking performance~\cite{pelrine-etal-2023-towards}. Whether this ordering advantage transfers to MDLMs, which are not constrained to generate tokens left-to-right, remains an open question. To investigate, we prompt each model to produce the verdict ($V$) and justification ($J$) in both possible orders (Table~\ref{tab:baseline-order-performance}).

\input{tables/baseline-order-performance}

All three models exhibit minimal sensitivity to output ordering, though LLaMA 3.1 performs significantly worse overall than both Qwen3-8B and LLaDA-8B. Examining LLaDA's token trajectories offers insight into why ordering has little effect: regardless of the prompted sequence ($J \xrightarrow{} V$ or $V \xrightarrow{} J$), the model resolves the verdict within the first few diffusion steps, well before the justification is complete. In other words, LLaDA determines its answer immediately and fills in the reasoning afterward. We attribute this behavior to the relatively low complexity of binary fact verification in this setting, which allows the model to commit to a verdict without extended deliberation.

\input{tables/deliberation}

\subsection{Delayed Verdict Unmasking}\label{sec:delayed-unmasking}
The preceding result raises a natural follow-up question: does forcing the model to deliberate longer improve accuracy? To test this, we prevent the verdict token from being unmasked until a specified percentage of justification tokens have been revealed (Table~\ref{tab:deliberation}). Contrary to expectation, this intervention harms performance, reducing accuracy from 86.2\% to 71.9\%. Interestingly, we notice no significant degredation in performance until after 25\% deliberation.

Inspecting the verdict trajectories reveals the underlying cause: the model frequently arrives at the correct verdict early in the diffusion process but reverses its prediction as additional justification tokens are generated. We refer to this phenomenon as refinement drift. As justification tokens accumulate, they introduce local inconsistencies that progressively conflict with the model's initially correct global assessment, ultimately causing it to override its own earlier---and more accurate---prediction. We also found the stability between 0--25\% deliberation also aligns with the generation of the JSON structure of the output which typically occurs within the first 10-16 tokens.

\subsection{Interventional Faithfulness Analysis}\label{sec:faithfulness}
To probe the causal relationship between justification and verdict, we exploit a distinctive property of MDLMs: the ability to condition on arbitrary subsets of tokens. This enables two complementary interventions---one testing whether the model will rationalize a forced verdict, and another testing whether it genuinely relies on justifications when predicting verdicts.

\paragraph{Integrity Test.}
We hard-code an incorrect verdict and allow the model to infill the surrounding justification. Using Qwen3-30B-A3B as a judge (categories detailed in Appendix~\ref{app:llm-judge-categories}), we find that in 44\% of cases the model maintains logical integrity, generating a justification that contradicts the forced verdict and instead supports the correct answer. In the remaining cases, the model rationalizes the incorrect verdict, either through logical errors (37\%) or by hallucinating facts not present in the evidence (13\%).\footnote{The remaining 6\% were classified as ``other.''}

\paragraph{Reliance Test.}
We next ask the converse question: does LLaDA genuinely rely on its own justifications when predicting a verdict? To test this, we take the corrupted justifications produced in the Integrity Test and hard-code them as context, forcing the model to infill only the verdict token. To prevent verdict leakage through the justification text, we mask tokens corresponding to label-indicative words and phrases such as \textit{true}, \textit{false}, \textit{supported}, and \textit{unsupported} (full list in Appendix~\ref{app:masked-justification-tokens}).

When conditioned on corrupted justifications, accuracy drops sharply to 57.3\%. This decline is driven primarily by a near-total failure on supported claims (17.2\% accuracy), while performance on refuted claims degrades more modestly (73.4\% accuracy); we analyze this asymmetry in Appendix~\ref{app:asymmetry-analysis}. In contrast, when the model is provided with high-quality ground-truth justifications from the AVeriTeC dataset, accuracy reaches 97.1\%.

\paragraph{Implications.}
This disparity demonstrates that the model's verdict is strongly causally dependent on the quality of its reasoning trace. Crucially, this dependence does not contradict the early verdict convergence observed in Section~\ref{sec:justification-ordering}---rather, it reveals the mechanism underlying refinement drift (Section~\ref{sec:delayed-unmasking}). The model commits to a verdict within the first few diffusion steps, then generates justification tokens that it subsequently conditions on. When these tokens contain errors, they exert pressure in the wrong direction, gradually overriding the initially correct assessment.

%% file: tables/baseline-order-performance.tex
\begin{table}[t]
    \centering
    \begin{tabularx}{\linewidth}{Xcc}
        \toprule
        \textbf{Model} & \textbf{Output Order} & \textbf{Acc (\%)} \\ 
        \midrule
        LLaMA 3.1 (8B)      & $V \xrightarrow{} J$  & 69.7 \\
        LLaDA-8B            & $V \xrightarrow{} J$  & 86.2 \\
        Qwen3-8B            & $V \xrightarrow{} J$  & {89.5} \\
        \midrule
        LLaMA 3.1 (8B)      & $J \xrightarrow{} V$  & 72.5 \\
        Qwen3-8B            & $J \xrightarrow{} V$  & 88.0 \\
        LLaDA-8B            & $J \xrightarrow{} V$  & {89.0} \\
        \bottomrule
    \end{tabularx}
    \caption{Performance of baseline models on AVeriTeC development set. Order indicates the order the justification ($J$) and verdict ($V$) are generated in.}
    \label{tab:baseline-order-performance}
\end{table}

%% file: tables/deliberation.tex
\begin{table}[t]
    \centering
    \begin{tabularx}{\linewidth}{Xcc}
        \toprule
        \textbf{Model} & \textbf{Deliberation (\%)} & \textbf{Acc (\%)} \\ 
        \midrule
        LLaDA-8B & 0  & 86.2 \\
        LLaDA-8B & 25 & 86.6 \\
        LLaDA-8B & 50 & 79.2 \\
        LLaDA-8B & 75 & 73.8 \\
        LLaDA-8B & 90 & 71.9 \\
        \bottomrule
    \end{tabularx}
    \caption{Performance impact of forced deliberation on LLaDA. Deliberation \% is calculated based on the original span of 64 reasoning tokens.}
    \label{tab:deliberation}
\end{table}

%% file: sections/5-conclusion.tex
\section{Conclusion}
In this work, we investigated how LLaDA-8B handles fact verification on the AVeriTeC dataset. We find that the model achieves competitive performance and is robust to output ordering, but consistently resolves the verdict within the first few diffusion steps---even when instructed to generate justifications first. Forcing extended deliberation through delayed verdict unmasking proves counterproductive, degrading accuracy from 86.2\% to 71.9\% as accumulating justification tokens override initially correct predictions. Our interventional analyses reveal the underlying mechanism: LLaDA's verdicts are causally dependent on justification quality. When forced to justify incorrect verdicts, the model maintains logical integrity in only 44\% of cases, and conditioning on these corrupted justifications drops accuracy to 57.3\%---compared to 97.1\% with ground-truth justifications. These findings suggest that for current MDLMs, extended deliberation under constrained decoding risks diluting accurate early predictions with noise introduced during justification generation, though future work is needed to determine whether this pattern generalizes to other reasoning tasks and model scales.

%% file: sections/appendix.tex
\input{tables/examples}
\section{Appendix}

\subsection{Reproducibility}
\label{app:reproducibility}
All results are based on single runs. LLaDA's unmasking process is deterministic when unmasking one token at a time (selecting the highest-probability token at each step), producing identical outputs across runs. For LLaMA 3.1 and Qwen3, we used single runs with default generation parameters, as these models serve as baselines and are not the focus of our analysis. GitHub Copilot was used as an autocomplete tool during development.

\subsection{Reasoning Trace Categories}
\label{app:llm-judge-categories}
To categorize how the model handles incorrect forced verdicts in the Integrity Test, we manually reviewed approximately 50 justifications and identified the following error categories, which were then used as instructions for the LLM judge (Qwen3-30B-A3B):
\begin{itemize}
    \item Logical error: The justification contains reasoning errors that lead to an incorrect conclusion.
    \item Cherrypicking details: The justification selectively uses evidence that supports the incorrect verdict while ignoring evidence that contradicts it.
    \item Verdict-justification mismatch: The justification does not logically support the given verdict (i.e., the justification suggests a different verdict).
    \item Factual hallucination to support wrong verdict: The justification includes evidence which is not explicitly given that are used to support the incorrect verdict.
    \item Other (describe in a few words): The justification contains an error that does not fit into the above categories.
\end{itemize}

\subsection{Masked Justification Tokens}
\label{app:masked-justification-tokens}

During the Reliance Test, we masked all tokens corresponding to the following label-indicative words and phrases to prevent verdict leakage through the justification text: \textit{true}, \textit{false}, \textit{True}, \textit{False}, \textit{TRUE}, \textit{FALSE}, \textit{no evidence}, \textit{unsupported}, \textit{refuted}, \textit{supported}, \textit{support}, \textit{refute}, \textit{inconsistent}, \textit{consistent}, \textit{accurately}, \textit{inaccurately}, \textit{incorrectly}, \textit{correctly}, \textit{impossible to verify}, \textit{impossible to determine}.

\subsection{Supported/Refuted Accuracy Asymmetry}
\label{app:asymmetry-analysis}

In the Reliance Test, accuracy on supported claims (17.2\%) was substantially lower than on refuted claims (73.4\%). We attribute this asymmetry to the linguistic properties of fact-checking language. After masking verdict-leaking tokens, refuted justifications retain unambiguous signals such as ``hoax,'' ``satire,'' and ``fake news,'' while supporting justifications lose most of their discriminative content---words like ``confirmed'' become ambiguous without surrounding context. This asymmetry is further compounded by the inherent directness of refutation language compared to the hedged phrasing typical of support (e.g., ``evidence suggests,'' ``although not confirmed'').

The model's agreement rates confirm this pattern: it follows refuted justifications 85.6\% of the time but follows supported justifications only 26.6\% of the time. When positive signals are weak, the model appears to fall back on a prior---likely acquired during pretraining---that claims under fact-checking scrutiny tend to be false.

\subsection{Example Predictions with Corrupted Justifications}
\label{app:example-predictions}
Table~\ref{tab:example-predictions} shows examples from the Reliance Test where corrupted justifications misled the model.

%% file: tables/examples.tex
\begin{table*}[th]
\centering
\small
\begin{tabularx}{\textwidth}{p{2.8cm}p{4.5cm}ccp{5.5cm}}
\toprule
\textbf{Claim} & \textbf{Evidence} & \textbf{Truth} & \textbf{Pred.} & \textbf{Generated Explanation \& Error} \\
\midrule
Amy Coney Barrett was confirmed as Supreme Court Justice on October 26, 2020.
& \textbf{Q:} Is Amy Coney Barrett confirmed as Supreme Court Justice? \textbf{A:} Yes.
& Supp. & Ref.
& \textcolor{red}{``...the date of confirmation is incorrect. She was confirmed on September 26, 2020.''} \newline \textit{Hallucinates wrong date; evidence has no dates.} \\
\midrule
Jammu and Kashmir was removed from the UN list of `unresolved disputes'.
& \textbf{Q:} Was J\&K an item on the Security Council agenda? \textbf{A:} ``J\&K has never been an item on the agenda... it is listed in a separate category.''
& Ref. & Supp.
& \textcolor{red}{``The claim is true because the dispute has never been an item on the agenda... indicating it has been removed.''} \newline \textit{Conflates ``never listed'' with ``removed.''} \\
\bottomrule
\end{tabularx}
\caption{Example predictions from the Reliance Test showing how corrupted justifications mislead the model. Supp.\ = Supported, Ref.\ = Refuted.}
\label{tab:example-predictions}
\end{table*}